\pdfoutput=1

\documentclass[10pt,twocolumn,letterpaper]{article}

\usepackage[pagenumbers]{actusarxiv}

\usepackage{xcolor}

\usepackage{amsthm,bm,mathtools}
\usepackage[most]{tcolorbox}
\usepackage{algorithm,algpseudocode}
\usepackage{multirow}
\usepackage{longtable}
\usepackage{colortbl}%
\definecolor{uwmblue}{HTML}{1F5BA8}
\definecolor{uwmblueL}{HTML}{C6DAF1}
\definecolor{uwmblueXL}{HTML}{EBF2FA}
\definecolor{uwmscarlet}{HTML}{BA0C2F}
\definecolor{uwmgray}{HTML}{646A6E}
\definecolor{uwmgrayL}{HTML}{868E92}
\definecolor{uwmgreen}{HTML}{23795A}%
\usepackage{xspace}

\newcommand{\R}{\mathbb{R}}
\newcommand{\E}{\mathbb{E}}
\newcommand{\Prob}{\mathbb{P}}

\newcommand{\Cov}{\operatorname{Cov}}
\newcommand{\tr}{\operatorname{tr}}

\newcommand{\abox}{\mathcal{A}}%
\newcommand{\Filt}{\mathcal{F}}%

\newcommand{\Tf}[1]{T(#1)}%

\newcommand{\norm}[1]{\lVert #1\rVert}
\newcommand{\Fnorm}[1]{\lVert #1\rVert_{F}}%
\newcommand{\Info}{\mathcal{I}}%
\newcommand{\Ieff}{\mathcal{I}_{\mathrm{eff}}}
\newcommand{\lameff}{\lambda_{\mathrm{eff}}}
\newcommand{\Rao}{\Lambda}
\newcommand{\GLR}{\Lambda^{\mathrm{GLR}}}
\newcommand{\ncx}[2]{\chi^2_{#1}(#2)}
\newcommand{\cx}[1]{\chi^2_{#1}}
\newcommand{\thz}{\theta_0}
\newcommand{\dhat}{\hat\delta}
\newcommand{\Snom}{S_{\mathrm{nom}}}
\newcommand{\Sfro}{S_{\mathrm{f}}}
\newcommand{\Sad}{\hat S}
\newcommand{\Sor}[1]{S^{\star}_{#1}}
\newcommand{\Lbox}{L_{\mathrm{box}}}
\newcommand{\Llat}{L_{\mathrm{lat}}}
\newcommand{\Lest}{L_{\mathrm{est}}}

\newcommand{\actus}{\texorpdfstring{\textup{\textbf{\textsc{\color{uwmblue}Actus}}}}{ACTUS}\xspace}

\newtcbox{\RQtag}{on line, nobeforeafter, tcbox raise base, arc=1.6pt, boxrule=0pt, boxsep=0pt,
  left=1.7pt, right=1.7pt, top=0.9pt, bottom=0.5pt, colback=uwmscarlet, opacityback=0.13,
  colframe=uwmscarlet, opacityframe=0, fontupper=\sffamily\bfseries\footnotesize\color{uwmscarlet!85!black}}
\newcommand{\RQ}[2]{\RQtag{RQ#1}\ \textbf{#2}}
\newcommand{\RQref}[1]{\RQtag{RQ#1}}

\newtcolorbox[use counter=sptheorem]{theorembox}[2][]{%
  enhanced, colback=red!3!white, colframe=red!60!black,
  coltitle=white, fonttitle=\bfseries\small,
  title={Theorem~\thetcbcounter: #2}, boxrule=0.6pt, arc=1.5pt,
  left=4pt, right=4pt, top=3pt, bottom=3pt, before skip=6pt, after skip=6pt, #1}

\newtcolorbox[use counter=spdefinition]{definitionbox}[2][]{%
  enhanced, colback=green!3!white, colframe=green!40!black,
  coltitle=white, fonttitle=\bfseries\small,
  title={Definition~\thetcbcounter: #2}, boxrule=0.6pt, arc=1.5pt,
  left=4pt, right=4pt, top=3pt, bottom=3pt, before skip=6pt, after skip=6pt, #1}

\newtcolorbox[use counter=spproposition]{propositionbox}[2][]{%
  enhanced, colback=red!3!white, colframe=red!60!black,
  coltitle=white, fonttitle=\bfseries\small,
  title={Proposition~\thetcbcounter: #2}, boxrule=0.6pt, arc=1.5pt,
  left=4pt, right=4pt, top=3pt, bottom=3pt, before skip=6pt, after skip=6pt, #1}
\newtcolorbox{takeawaybox}[1]{%
  enhanced, colback=gray!6!white, colframe=gray!55!black,
  coltitle=white, fonttitle=\bfseries\small, title={#1},
  boxrule=0.6pt, arc=1.5pt, left=4pt, right=4pt, top=3pt, bottom=3pt,
  before skip=6pt, after skip=6pt}
\theoremstyle{remark}
\newtheorem{assumption}{Assumption}

\theoremstyle{plain}

\theoremstyle{remark}

\newcommand{\uwmSecFmt}[1]{\crefformat{#1}{\S##2##1##3}%
  \crefmultiformat{#1}{\S\S##2##1##3}{ and~##2##1##3}{, ##2##1##3}{ and~##2##1##3}%
  \crefrangeformat{#1}{\S\S##3##1##4--##5##2##6}}
\AtBeginDocument{%
  \crefname{appendix}{Sec.}{Secs.}\Crefname{appendix}{Section}{Sections}%
  \crefname{subappendix}{Sec.}{Secs.}\Crefname{subappendix}{Section}{Sections}%
  \crefname{subsubappendix}{Sec.}{Secs.}\Crefname{subsubappendix}{Section}{Sections}%
  \uwmSecFmt{section}\uwmSecFmt{subsection}\uwmSecFmt{subsubsection}%
  \uwmSecFmt{appendix}\uwmSecFmt{subappendix}\uwmSecFmt{subsubappendix}}

\definecolor{linkblue}{rgb}{0.21,0.49,0.74}
\usepackage[pagebackref,colorlinks,allcolors=linkblue]{hyperref}
\hypersetup{pdftitle={How Uncertain Is My Action? Self-Correcting Visual Planning with Frozen World Models},
  pdfauthor={Siqing Li, Luming Yang, Yijia Jiao, Junze Deng, Alper Yilmaz},
  pdfsubject={}, pdfkeywords={}, pdfcreator={LaTeX}}

\input{supp_labels}

\title{\includegraphics[height=0.8em,trim=10.9 17.7 46.9 18.2,clip]{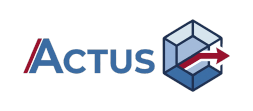}\hspace{0.15em}\raisebox{-0.23em}{\includegraphics[height=1.26em,trim=77.5 9.1 10.76 9.1,clip]{figures/actus_logo.pdf}}\hspace{0.4em}How Uncertain Is My Action?\\Self-Correcting Visual Planning with Frozen World Models}

\author{Siqing Li, Luming Yang\textsuperscript{\dag}, Yijia Jiao, Junze Deng, Alper Yilmaz\\
The Ohio State University}

\begin{document}
\twocolumn[{%
\renewcommand\twocolumn[1][]{#1}%
\maketitle
\begin{center}
\centering
\includegraphics[width=\textwidth]{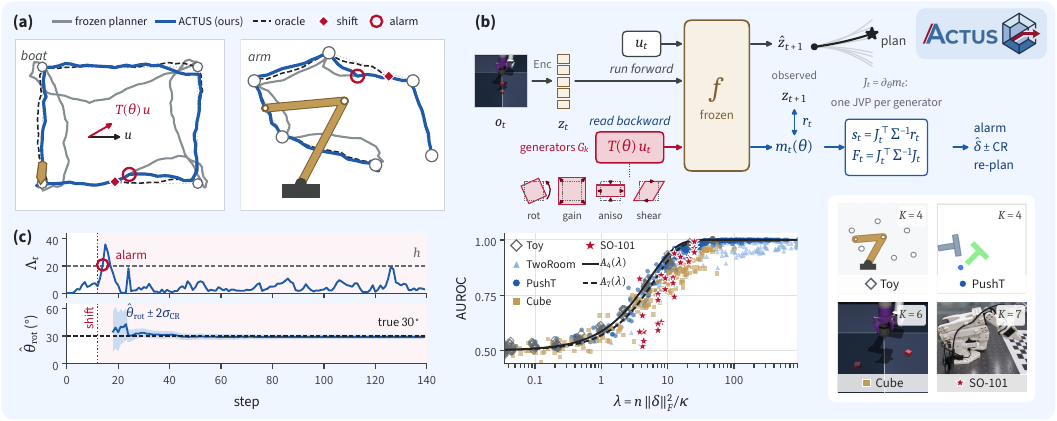}
\vspace{-0.6em}
\captionof{figure}{\textbf{Reading a frozen world model backwards.}
\textbf{(a)} A toy boat and a two-link arm under an unknown actuation shift
$\Tf\theta$: the frozen planner (grey) drifts, \actus (Action Uncertainty Self-correction, our method, blue) recovers after
its alarm.
\textbf{(b)} The frozen predictor $f$ is run forward to plan and backward
to read the shift score $s_t$ and its information $F_t$.
\textbf{(c)} Left: Rao statistic and rotation estimate with its
Cram\'er--Rao band on the boat. Middle: window AUROC against
$\lambda=n\Fnorm\delta^2/\kappa$ in all settings; lines: Prop.~2. Right: the four settings.}
\label{fig:teaser}
\end{center}
\vspace{0.4em}

}]

\begin{NoHyper}
\renewcommand\thefootnote{}%
\footnotetext{All authors contributed equally.}%
\footnotetext{\textsuperscript{\dag}Corresponding author: \texttt{yang.7670@osu.edu}. Photogrammetric Computer Vision Laboratory (PCVLab).}%
\end{NoHyper}

\begin{abstract}

A frozen latent world model combined with sampling-based model predictive control is a visual planner. At deployment, the executed action can depart from the commanded one, and the frozen model has no way to notice. We show that the model can be run in reverse: for any hypothesized actuation shift, it predicts what it should have observed under that shift, and comparing this prediction with what was actually observed quantifies the shift and lets the planner correct itself. Unlike the residual energy, which mixes the shift with the model's own error, the score keeps only the part of the residual that a shift can produce. This shift score detects the shift with a locally optimal test, estimates it with Cramér–Rao error bars, names the dominant generator, and compensates the planner once an alarm fires. On PushT, TwoRoom and OGBench-Cube with a frozen latent world model, measured detection follows the prediction to within 0.05 AUROC. Self-correction raises success from 0.650 to 0.862 on PushT and from 0.378 to 0.432 on the harder Cube task, levels with an oracle that is told the true shift at the same block, better than five common test-time adaptation methods. On a real robotic arm, the score identifies a base-joint offset that no joint encoder registers and reduces the reach error from 36 mm to 3--5 mm, and under shifts drawn at random and sealed before the session, estimates recover 92--94\% of the error reduction achieved with the true shift.
The broader implication is that a frozen world model can act as a filter by predicting the future, testing action-interface uncertainty and validity, and finally correcting failures.

\end{abstract}

\section{Introduction}
\label{sec:intro}

Visual planning often pairs a frozen latent world model (WM) trained on pixels with sampling-based model predictive control (MPC). The planner rolls out candidate action sequences, selects those predicted to reach the goal, and executes the first action block \cite{zhou2025dinowm,maes2026lewm,assran2025vjepa2,sobal2025pldm}. This uses the model only in the forward direction, predicting future states from observations and commanded actions. At deployment, however, a miscalibrated axis or a slipped joint can silently transform every command. We call this an \emph{actuation shift}. The frozen model still assumes commands are executed as given, so its preferred plan can drive the robot away from the goal (\cref{fig:teaser}a). In other words, the model has no notion of execution uncertainty; it assumes every command is executed as given and cannot detect when it is not.

Existing remedies act on the model's prediction error \cite{domberg2025wmanomaly}, and test-time adaptation reduces it by updating weights or auxiliary modules \cite{wang2026adajepa,mete2026optta,zhang2026jepattt,soni2026sandwich,soni2026alra}, without separating an actuation shift from the model's own error.

Our starting point is that a frozen world model is a \emph{counterfactual simulator of its own actions} \cite{pearl2009,buesing2019woulda,bear2023cwm}. For each hypothesized shift $\theta$, the frozen model predicts the next latent as if the command were executed through $\Tf\theta$. Comparing this prediction with the observed next latent reveals which hypothesis best explains the transition, without further training. Inverting the simulator over a Lie group \cite{sola2018} of action transformations turns the question ``\textit{is the model wrong?}'' into an estimation problem amenable to the statistical local approach to fault detection \cite{benveniste1987,basseville1998}. This perspective motivates a filter-like diagnostic: calibrate the prediction residual and test whether it indicates an actuation shift.

Building on this view, we propose \textbf{Act}ion \textbf{U}ncertainty \textbf{S}elf-correction (\actus), which lets a frozen world model quantify its execution uncertainty and correct actuation shifts without any weight update. 
At its core is the \emph{shift score}, which projects the model's residual onto the directions in which each generator would move the prediction, given by a Jacobian--vector product (JVP) of the frozen model along that
generator. Without a shift, its covariance is the Fisher information (\cref{fig:teaser}b).
From this single statistic, \actus detects the shift with a locally optimal Rao test, estimates it by Gauss--Newton (GN) with Cram\'er--Rao error bars, and attributes it to the dominant generator. Once an alarm fires, it compensates for the estimated shift by replanning through its inverse, enabling self-correcting visual planning without updating the world model. Calibration on nominal rollouts accounts for model error and variation across sessions, allowing us to predict detection and estimation performance before a shift occurs.

We conduct experiments with a frozen LeWM \cite{maes2026lewm} on three standard visual planning benchmarks, PushT \cite{chi2023diffusionpolicy}, TwoRoom \cite{sobal2025pldm} and OGBench-Cube \cite{park2025ogbench}, and demonstrate that \actus detects and corrects actuation shifts without any weight updates. As shown in \cref{fig:teaser}c, measured detection follows the predicted law across all $254$ shifts. Compensation brings the frozen planner close to an \emph{oracle} told the true shift at the same block, outperforming all baselines. \actus further transfers unchanged to DINO-WM \cite{zhou2025dinowm} and remains effective on a real SO-101 robotic arm.

To summarize, our main contributions are:
\begin{enumerate}[leftmargin=*,itemsep=1pt,topsep=2pt]
\item \textbf{Inverse counterfactual formulation.}
We formulate action shifts as Lie-group transformations of commands. We show that the residual energy is insensitive to shifts at first order and derive a \emph{shift score} computed with one JVP per generator, whose detection and estimation performance can be predicted from nominal data (\cref{sec:prelim} and \cref{sec:method}).

\item \textbf{A self-correcting framework.}
We propose \actus, which combines calibrated Rao detection, GN estimation with uncertainty quantification, generator attribution, and alarm-gated inverse compensation to correct actuation shifts without updating model weights (\cref{sec:method}).

\item \textbf{Empirical validation.}
Across $254$ shifts, detection AUROC matches predictions within $0.05$ and exceeds residual energy by up to $0.38$. Compensation raises success significantly, matching the same-block oracle and outperforming the adaptation baselines. \actus also transfers to DINO-WM without retuning (\cref{sec:exp}).

\item \textbf{Real-world Application.} We validate \actus on a real robot arm, detecting a base-joint offset invisible to joint readings and reducing reach error from $36$--$38$\,mm to $3.0$--$5.5$\,mm, matching repeat noise (\cref{sec:exp-real}).
\end{enumerate}

\section{Related work}
\label{sec:related}

\paragraph{Latent world models and visual planning.}
Planning through a learned latent dynamics model goes back to
\cite{ha2018worldmodels,hafner2019planet,hafner2020dreamer}. The recent
generation trains the model on pixels with a joint-embedding predictive architecture and
plans with sampling-based MPC against a goal image
\cite{sobal2025pldm,assran2025vjepa2,maes2026lewm}; 
The WM toolkit \cite{maes2026swm} provides our environments.
These models are run forward, and
recent probes ask what else a trained model knows about its actions
\cite{vasilev2026geometry,wang2026groupactions,nam2026}. We ask inversely: what type of action-input transformation explains the model observation?

\paragraph{Test-time adaptation of world models.}
Test-time adaptation updates a model on test data
\cite{sun2020ttt,wang2021tent,nagabandi2019adapt}. For latent world models
in the MPC loop, AdaJEPA \cite{wang2026adajepa}, OpTTA \cite{mete2026optta},
and JEPA-TTT \cite{zhang2026jepattt} update the predictor;
Sandwich-Residuals and ALRA learn small residual modules
\cite{soni2026sandwich,soni2026alra}. They assume nothing about the form of
the shift and offer no calibration. We keep the model frozen, estimate
a shift with a false-alarm guarantee and Fisher error
bars, and compare against them under one planner.

\paragraph{Statistical fault detection.}
Our detector builds on the Rao score test \cite{rao1948}, Le~Cam's local
asymptotics \cite{lecam1986} and the statistical local approach to fault
detection and isolation \cite{benveniste1987,basseville1998}, along with the GLR
change detection of \cite{lorden1971,lai1998,basseville1993}.
Innovation-based diagnosis tests whether a filter's innovation is white
\cite{kalman1960,mehra1971,willsky1976}. The model is given and
differentiated with respect to its own parameters. Here it is a frozen
learned simulator, differentiated with respect to a group acting on its
input, with its error as the noise law.
\S\ref{app:related} extends to action transformation,
adaptive control, model-based diagnosis, and failure detection for learned
controllers.

\section{Preliminaries}
\label{sec:prelim}

\subsection{Frozen world models and actuation shifts}
\label{sec:prelim:wm}
\label{sec:prelim:shift}

A latent world model consists of an encoder $z_t=\mathrm{Enc}(o_t)\in\R^D$
and a predictor $f$ that maps the recent latents $z_{\le t}$ and an action
block $a_t$ to the next latent, $\hat z_{t+1}=f(z_{\le t},a_t)$. A block
is a sequence of $F$ sub-actions $u_{t,1},\dots,u_{t,F}\in\R^m$
(end-effector increments), each clipped to the command box
$\abox=[-1,1]^m$ by the environment. The planner is sampling-based MPC
(CEM, \cite{rubinstein1999}) that executes the first block of the best
sequence. $\mathrm{Enc}$ and $f$ are frozen: we use LeWM
\cite{maes2026lewm} and DINO-WM \cite{zhou2025dinowm} as released. Between
the planner and the environment sits an actuator that may not do what it
is told.

\begin{assumption}[Structured actuation shift]\label{ass:shift}
The executed sub-action is $\tilde u_{t,j}=\mathcal P(\Tf\theta\,u_{t,j})$,
with $\mathcal P$ the clipping and
\begin{equation}
  \Tf{\theta}=\exp\Big(\sum_{k=1}^{K}\theta_kG_k\Big),\qquad\theta\in\R^K,
  \label{eq:lie}
\end{equation}
where the \emph{shift generators} $\{G_k\}$ are linearly independent
\cite{sola2018}. The coordinates
$\theta$ are unknown and may drift slowly.
\end{assumption}

For planar increments, we take all of $\mathfrak{gl}(2)$ (rotation,
isotropic gain, anisotropic gain, shear; $K=4$), and for spatial increments,
the three axis gains with $\mathfrak{so}(3)$ ($K=6$), a span not closed
under the bracket, so composition holds there to first order
(\S\ref{app:lie}). Rotations are in degrees and gains are factors.

\subsection{Counterfactual prediction and the residual}
\label{sec:prelim:residual}

A frozen world model answers, at no cost, a counterfactual question: for
any shift hypothesis $\theta$,
\begin{equation}
  m_t(\theta):=f\big(z_{\le t},A_\theta(a_t)\big),\quad
  A_\theta(a_t)=\big(\mathcal P(\Tf{\theta}u_{t,j})\big)_{j\le F},
  \label{eq:cf}
\end{equation}
is the latent it would have predicted had the commanded block been executed
through $\Tf\theta$. The factual observation $z_{t+1}$ defines a residual
and its first-order model,
\begin{equation}
  r_t:=z_{t+1}-m_t(\thz),\qquad r_t\approx J_t\,\delta+\varepsilon_t .
  \label{eq:residual}
\end{equation}
Here $\thz$ is the \emph{zero point}, the shift the frozen model implicitly
assumes on nominal data, and $\delta$ is the shift relative to it,
$\Tf\theta=\Tf\thz\Tf\delta$. The Jacobian
$J_t=\partial m_t/\partial\delta|_0\in\R^{D\times K}$ costs one forward-mode
JVP of $f$ per generator, and $\varepsilon_t$ is the model's own
prediction error.

\begin{assumption}[Calibrated nominal law]\label{ass:nominal}
Under nominal actuation,
$J_t^{\top}\Sigma_t^{-1}\E[\varepsilon_t\,|\,\Filt_t]=0$ (zero mean
along $J_t$) and
$\Cov(\varepsilon_t\,|\,\Filt_t)=\Sigma_t=\Sigma_r(a_t)$,
where $\Filt_t$ is the history up to the choice of $a_t$. The zero point
$\thz$ and the residual law $\Sigma_r(\cdot)$ are estimated from
nominal rollouts and then frozen.
\end{assumption}

This is the setting of the statistical local approach to fault detection
\cite{basseville1998,benveniste1987}, with the physical model replaced by a
learned one. The three questions of the paper become statements about shift
$\delta$: \emph{detection} (is $\delta\ne0$?), \emph{estimation and
compensation} (replace $a$ by $\Tf\dhat^{-1}a$), and \emph{attribution}
(which generator dominates $\delta$). The structure of
\cref{eq:lie} and Assumption~\ref{ass:nominal} is what makes $\delta$
identifiable from factual data \cite{pearl2009} (\S\ref{app:theory-notes}).

\subsection{Limits of the residual energy}
\label{sec:prelim:norm}

\begin{figure}[t]
\centering
\includegraphics[width=\columnwidth]{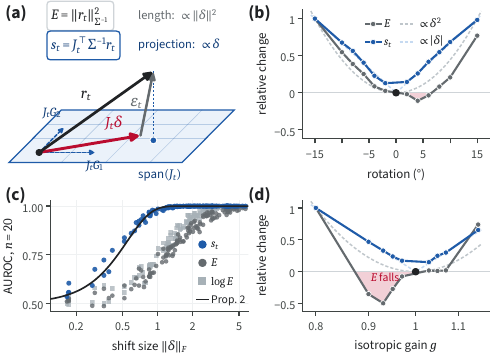}
\caption{\textbf{Energy against score (Prop.~1).} \textbf{(a)} a
shift moves the residual $r_t$ only inside the span of $J_t$ (blue), the
model error $\varepsilon_t$ in all $D$ directions. \textbf{(b,d)} change of the whitened energy $E$ (grey) and of the score
(blue) on PushT along two generator axes. \textbf{(c)} AUROC against the Fisher norm, $124$ PushT shifts; line: Prop.~2.}
\label{fig:geometry}
\end{figure}

The obvious statistic is the model's surprise, the whitened residual
energy $E:=\sum_{t\in\mathcal W}r_t^{\top}\Sigma_t^{-1}r_t$ over a window
$\mathcal W$ of $n$ transitions, the statistic of innovation-based
diagnosis \cite{mehra1971,willsky1976} and of prediction-error monitors.
Its nominal counterpart is
$E_0:=\sum_{t\in\mathcal W}\varepsilon_t^{\top}\Sigma_t^{-1}\varepsilon_t$,
with $\E[E_0]=nD$ for a calibrated model.
Let $F_t:=J_t^{\top}\Sigma_t^{-1}J_t$ be the per-transition Fisher
information about $\delta$, $\bar F:=\tfrac1n\sum_{t\in\mathcal W}F_t$ its
window mean, and $\Fnorm\delta:=(\delta^{\top}\bar F\delta)^{1/2}$ the
\emph{Fisher norm}, the unit in which the paper measures shifts.

\begin{propositionbox}{Residual energy confounds the shift with the model's own error}
Under Assumptions~\ref{ass:shift}--\ref{ass:nominal} and the linearization
\eqref{eq:residual}, with $E$ computed from the raw prediction $m_t(0)$, as
a prediction-error monitor would,
\begin{align}
  \E[E]-\E[E_0]
  &=\underbrace{\sum_{t}\norm{J_t\thz}^2_{\Sigma_t^{-1}}}_{\text{model bias}}
   +\underbrace{2n\,\thz^{\top}\bar F\delta}_{\text{bias}\times\text{shift}}
   \notag\\
  &\quad+\underbrace{n\,h(\delta)}_{\text{residual scale}}
   +\underbrace{n\,\Fnorm\delta^2}_{\text{shift}},
  \label{eq:energy}
\end{align}
where $h(\delta)$, the change of the residual scale with the executed
motion, vanishes for isometric generators and is first order in the gain coordinates with the sign of the gain.
Only the last term grows with every shift, and it is quadratic: against
the shift score of \S\ref{sec:detect} the energy loses a factor
$\sqrt{2\Fnorm\delta^2/D}$ for small shifts. With the calibrated zero point
of \cref{eq:residual} in place of $m_t(0)$, only the first two terms vanish. (Proof in
\S\ref{app:norm}.)
\end{propositionbox}

The energy thus adds up effects it cannot separate (\cref{fig:geometry}).
The model's own bias inflates the null; a shift aligned with that bias
raises the surprise, while one that cancels it \emph{lowers} it, and a reduced
gain moves less. Therefore, the model errs less, although the actuator is faulty.
The shift's own contribution is second order because a norm averages the
displacement $J_t\delta$ over all $D$ latent directions, of which only $K$
carry it. The signal lies in the \emph{direction} of the residual within
the span of $J_t$, where a shift enters at first order with the
positive-definite coefficient $\bar F$ \cite{basseville1997}.

\section{\actus: self-correction from the shift score}
\label{sec:method}

\begin{figure*}[t]
\centering
\includegraphics[width=\textwidth]{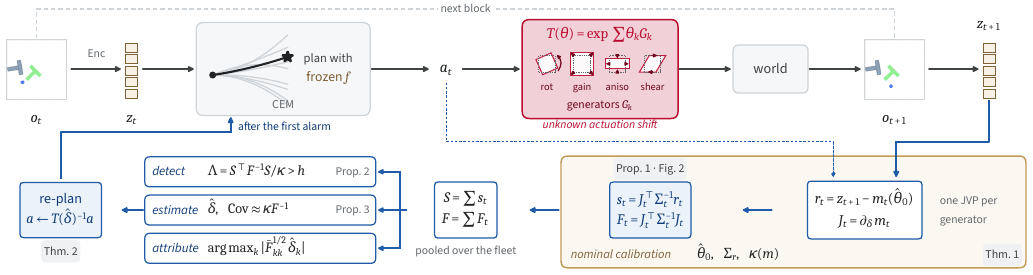}
\caption{\textbf{Overview of \actus.} The frozen model plans as
usual and the actuator applies the unknown shift $\Tf\theta$ (red). The residual and one JVP per generator give
$(s_t,F_t)$, pooled into $(S,F)$ for alarm, the estimate $\dhat$; after the first alarm the planner re-plans through
$\Tf{\dhat}^{-1}$.}
\label{fig:overview}
\end{figure*}

To address actuation shifts during visual planning, we introduce \actus,
which reads one $K$-dimensional statistic from the frozen model at
every block, the \emph{shift score}, and uses it to detect a shift
(\S\ref{sec:detect}), to estimate and compensate it
(\S\ref{sec:estimate}), and to name its dominant generator
(\S\ref{sec:attribute}). Each use comes with a quantity computed before
any shift is applied (an AUROC, a delay, an error bar, a recovered
success rate), which \S\ref{sec:exp} tests (\cref{fig:overview};
pseudocode: Algorithm~\ref{alg:sp}).

\begin{definitionbox}{Shift score and shift information}
For transition $t$,
\begin{equation}
  s_t:=J_t^{\top}\Sigma_t^{-1}r_t,\qquad F_t:=J_t^{\top}\Sigma_t^{-1}J_t,
  \label{eq:score}
\end{equation}
and for any set $\mathcal W$ of $n$ transitions (a window, an episode, or
a fleet of episodes sharing the actuator) $S=\sum_{t\in\mathcal W}s_t$,
$F=\sum_{t\in\mathcal W}F_t$. Under Assumption~\ref{ass:nominal} and
\cref{eq:residual}, $\E[s_t|\Filt_t]=F_t\delta$ and
$\Cov(s_t|\Filt_t)=F_t$: the score is a $K$-dimensional measurement of
$\delta$ whose own noise is its information.
\end{definitionbox}

The pair $(s_t,F_t)$ is the efficient score of the local approach
\cite{basseville1998,benveniste1987}, with $J_t$ being the derivative of a
learned pixel-to-latent simulator with respect to a group acting on its
action input and $\Sigma_t$ the simulator's own calibrated error.

\paragraph{Linearization.}
Two expansions enter \cref{eq:residual}: of the shift map $\Tf\delta$ in
$\delta$ and of the frozen predictor $f$ in the action. Working at first
order in a local perturbation and re-linearizing at the current estimate
is the standard step of estimation on manifolds \cite{barfoot2017,sola2018}
and the premise of the local approach \cite{benveniste1987}. The
statements below lean on it to different degrees. The level of the
detector needs no expansion: under Assumption~\ref{ass:nominal} the score
is centered at $\delta=0$ whatever $f$ does away from it, and the test
stays consistent for any shift that moves the residual along $J_t$ on
average. The detectability law is local by design. The estimator
re-linearizes, which removes the curvature of the exponential map exactly
and leaves only the fidelity of the frozen model's counterfactual, a
property of the model tested in \S\ref{app:linear}.

\subsection{Detection}
\label{sec:detect}

\paragraph{Rao test.}
Pooled over a window, the score is asymptotically Gaussian,
$S\approx\mathcal N(F\delta,F)$, which needs only the two moments of
Assumption~\ref{ass:nominal} (martingale central limit theorem,
Lemma~\ref{lem:lan}). On a frozen
model the nominal score is over-dispersed because the model's error is not
white, and we write $\Cov(S)\approx\kappa F$ with a scalar $\kappa\ge1$
fitted on nominal data:
\begin{equation}
  \Rao:=S^{\top}F^{-1}S/\kappa .
  \label{eq:rao}
\end{equation}

\begin{propositionbox}{Detectability law}
Under Assumptions~\ref{ass:shift}--\ref{ass:nominal}, for a window of $n$
transitions with mean information $\bar F$ and $\Cov(S)=\kappa F$,
\begin{equation}
  \Rao\ \sim\ \ncx{K}{\lambda},\qquad \lambda=\frac{n\,\Fnorm\delta^2}{\kappa},
  \label{eq:lambda}
\end{equation}
asymptotically, including when the planner chooses $a_t$ from the past.
Hence the level-$\alpha$ power and the window AUROC are universal
functions of $\lambda$ alone,
\begin{equation}
\begin{aligned}
  \beta_K(\lambda)&=\Prob\big(\ncx{K}{\lambda}>q_{K,1-\alpha}\big),\\
  A_K(\lambda)&\approx\Phi\Big(\tfrac{\lambda}{2\sqrt{K+\lambda}}\Big),
\end{aligned}
  \label{eq:auroc}
\end{equation}
and the test $\{\Rao>q_{K,1-\alpha}\}$ is locally asymptotically optimal
among invariant tests \cite{lecam1986,lehmann2005}. (\S\ref{app:rao}.)
\end{propositionbox}

In Fisher units, environments, generators and sizes collapse onto one
curve whose only constants are $K$ and $\kappa$, which makes the AUROC of
every shift in \S\ref{sec:exp} a prediction. Inverting \cref{eq:auroc},
the smallest shift visible at AUROC $a$ in $n$ transitions is
$\Fnorm\delta=\sqrt{\kappa\lambda_a/n}$, with $\lambda_{0.75}=3.6$ for
$K=4$ and $4.2$ for $K=6$; afterwards this is the \emph{detection threshold} of
\S\ref{sec:exp}.

\paragraph{Calibration against the model's own error.}
Part of the model's error lies in the span of $J_t$ and is constant over
an episode or a recording session. To the detector it looks like a shift,
which makes the Rao statistic over-dispersed, more so when episodes are
pooled. We model it with three variance components \cite{laird1982},
\begin{equation}
  s_i=F_i\big(\delta+\xi_e+b_s\big)+\eta_i,
  \label{eq:hier}
\end{equation}
with episode effect $\xi_e\sim\mathcal N(0,\Omega)$, session effect
$b_s\sim\mathcal N(0,\Omega_s)$ and step over-dispersion
$\Cov(\eta_i)=cF_i$, all estimated from nominal data.

\begin{theorembox}{Detection limits set by the model's own error}
Under \eqref{eq:hier}, for a fleet of $m$ equal episodes in one session:
\textbf{(i)} the Rao statistic $S^{\top}F^{-1}S$ has null mean
$K\kappa(m)$ with
$\kappa(m)=c+\tfrac1K\tr(\Omega F_e)+\tfrac mK\tr(\Omega_sF_e)$, linear in the
fleet size; whitening $S$ by its full covariance
$B_m=F(cF^{-1}+\tfrac1m\Omega+\Omega_s)F$ gives an exact $\cx K$ null,
and the scalar $\kappa(m)=\tr(F^{-1}B_m)/K$ matches it at the mean with a
heavier upper tail;
\textbf{(ii)} the information about $\delta$ carried by the session is
$\Ieff=(\Omega_s+P_s^{-1})^{-1}\preceq\Omega_s^{-1}$ with
$P_s=\sum_e(cF_e^{-1}+\Omega)^{-1}$, so \eqref{eq:lambda} holds with
$\lameff=\delta^{\top}\Ieff\delta\le\delta^{\top}\Omega_s^{-1}\delta$: a shift
smaller than the session variability cannot be detected within a session,
however large the fleet;
\textbf{(iii)} a nominal reference from the same session with precision
$P_0$ removes $b_s$, giving $\Ieff'=(P_s^{-1}+P_0^{-1})^{-1}$.
(Proof in \S\ref{app:nuisance}.)
\end{theorembox}

Part (i) sets the thresholds under pooling and explains why the nominal
$\Rao$ matches $\cx K$ at the center and exceeds it in the upper tail; part (ii) places shifts below
the session variability inside the nominal band; and part (iii) is the
same-session reference used on the real arm. All three constants are
fixed on nominal data before any shift is applied
(\S\ref{app:theory-notes}).

\paragraph{Streaming detection.}
To localize the onset, we run the change-point generalized likelihood
ratio (GLR)
$\GLR_n=\max_{k<n}(S_n-S_k)^{\top}(F_n-F_k)^{-1}(S_n-S_k)/\kappa$ with a
Bonferroni threshold $h$ \cite{basseville1993}. The information bound of
\cite{lorden1971,lai1998} gives the expected delay
\begin{equation}
  \E[\tau-\nu]\approx\inf\Big\{d:\ \delta^{\top}\big(\Info(\nu{+}d)-\Info(\nu)\big)\delta\ge\kappa h\Big\},
  \label{eq:delay}
\end{equation}
with $\Info(t)=\sum_{k\le t}F_k$ the accumulated information
(\S\ref{app:delay}). Detection is a matter of information rather than
of steps, and a fleet of $m$ episodes divides the delay by $m$. In closed
loop the GLR alarm switches compensation on.

\subsection{Estimation and compensation}
\label{sec:estimate}

\paragraph{Estimator.}
The estimate $\dhat$ is the shift under which the frozen model best
explains the transitions seen so far,
\begin{equation}
  \dhat:=\arg\min_{\delta}\ \sum_t\norm{r_t(\delta)}^2_{\Sigma_t^{-1}},
  \label{eq:estimator}
\end{equation}
where $r_t(\delta)$ is the residual \eqref{eq:residual} recomputed
through $\Tf\thz\Tf\delta$. Setting the gradient to zero
gives the \emph{score equation}
$\sum_tJ_t(\dhat)^{\top}\Sigma_t^{-1}r_t(\dhat)=0$ \cite{wedderburn1974}:
the pooled score $S$ of \eqref{eq:score}, recomputed at $\dhat$, vanishes.
Gauss--Newton (GN) solves it by iterating
$\delta\leftarrow\delta+F^{-1}S$ with $(S,F)$ recomputed at the current
iterate (Fisher scoring \cite{nocedal2006}); five iterations suffice in
every environment. For drifting shifts the same updates form an
information Kalman filter with process noise $q$, which coincides with GN
at $q=0$ \cite{bell1993}.

\begin{propositionbox}{Precision follows the same information}
Under Assumptions~\ref{ass:shift}--\ref{ass:nominal}, any estimator
unbiased on the working model satisfies
$\Cov(\dhat_t)\succeq\kappa\,\Info(t)^{-1}$ \cite{cramer1946,rao1945}, and
the re-linearized GN estimator attains this bound to first order, removing
the bias of the one-step estimate $F^{-1}S$ due to the curvature of the
exponential map. (\S\ref{app:estimate}.)
\end{propositionbox}

\paragraph{Compensation.}
Once the alarm has fired, every planned block is replaced by
$\Tf{\dhat}^{-1}a$, so that the executed sub-action equals the planned one
whenever $\Tf\dhat=\Tf\delta$ and the compensated command stays in the
box. Compensation is \emph{alarm-gated}, which leaves nominal success
untouched at the design false-alarm rate, and the plan in service is
discarded at the alarm (\emph{immediate replanning}), because what costs
success is the delay until the plan changes.

\begin{theorembox}{Recovery decomposition}
Let $\Snom$ be the nominal success rate, $\Sor k$ the success of a planner
that uses the \emph{true} shift from block $k$ (with $\Sor\infty$ the
frozen planner) and $\Sad$ the success of the self-correcting planner that
switches at its alarm block $\tau$ with estimate $\dhat_\tau$. Then
$\Snom-\Sad=\Lbox+\Llat+\Lest$ with
\begin{equation}
\begin{aligned}
  \Lbox&:=\Snom-\Sor0\le\Psi(\delta),\\
  \Llat&:=\Sor0-\E[\Sor\tau]\approx\textstyle\sum_k\Prob(\tau\ge k)(\Sor{k-1}-\Sor k),\\
  \Lest&:=\E[\Sor\tau]-\Sad\approx\tfrac12\kappa\,\E\tr\big(-H_\tau\Info(\tau)^{-1}\big),
\end{aligned}
  \label{eq:decomp}
\end{equation}
where $\Psi(\delta)$ is the probability that a nominal episode issues a
command $u$ with $\Tf\delta^{-1}u\notin\abox$, $\Prob(\tau\ge k)$ follows
from \eqref{eq:delay} and $H_\tau$ is the curvature of success in the
shift. 
\end{theorembox}

Thm. 2 proof is in \S\ref{app:recovery}. The three terms have separate causes: the command box, which no correction
of the actuator can undo ($\Lbox$); the blocks planned before the alarm,
each costing the drop of the delayed-oracle curve ($\Llat$); and the
estimation error, second order in the Cram\'er--Rao bound ($\Lest$).
\S\ref{sec:exp} measures each term separately.

\subsection{Attribution}
\label{sec:attribute}

With $\dhat$ and its covariance $\kappa\Info^{-1}$, naming the shift is a
one-line rule: scale each coordinate by its own information,
$\tilde\delta_k=\bar F_{kk}^{1/2}\dhat_k$, and report the generator with
the largest $|\tilde\delta_k|$. Its confusion probabilities follow from the
same Cram\'er--Rao bound, and generators with nearly parallel columns of
$J_t$, such as rotation and shear, are the pair that confuses
\cite{basseville2002} (Prop.~\ref{prop:confusion}). A disturbance outside
the family is flagged by the energy the $K$ score directions do not
explain, $E_\perp:=E-\kappa\Rao$ (\S\ref{app:gof}).

\section{Experiments}
\label{sec:exp}

Our experiments are designed to answer the following core research
questions (RQs):
\begin{itemize}[leftmargin=*,itemsep=1pt,topsep=2pt]
\item \RQ{1}{Detection.} Can \actus detect an actuation
shift in the frozen world model?
\item \RQ{2}{Estimation and compensation.} How much task success does
compensation return, compared with an oracle told the true shift and with
test-time adaptation?
\item \RQ{3}{Attribution.} Does \actus correctly name the dominant generator of an unknown shift?
\item \RQ{4}{Generalization.} Does the same procedure carry over to a
second world model without modification?
\item \RQ{5}{Real-world application.} Does it detect, estimate, and
compensate for shifts on a real robot arm?
\end{itemize}

\subsection{Experimental setup}
\label{sec:exp-setup}

\paragraph{Environments and world models.}
We use the released LeWM checkpoints \cite{maes2026lewm} for PushT \cite{chi2023diffusionpolicy}, TwoRoom \cite{sobal2025pldm} and OGBench-Cube \cite{park2025ogbench}, loaded strictly and never updated, and repeat the protocol on DINO-WM \cite{zhou2025dinowm}. PushT and TwoRoom are driven by planar increments with $K=4$ generators, and Cube by spatial increments with $K=6$. The planner is the CEM of the LeWM release. An episode lasts twelve blocks of five environment steps, and success is the environment's own criterion (\cref{app:protocol}).

\paragraph{Baselines.}
All methods share weights, planner, replanning schedule, start--goal pairs and shift draws, and every baseline is tuned on held-out shifts and pairs. They fall into four groups, described in full in \cref{app:baselines}.
\begin{itemize}[leftmargin=*,itemsep=0pt,topsep=1pt]

\item \emph{Reference planners.} Nominal runs without shift, Frozen ignores the shift, and Oracle@$k$ is given the ground-truth shift from block $k$ onward and serves as a reference.

\item \emph{Detection baselines.} The residual energy, its whitened form $E$, the innovation-mean $\chi^2$ test \cite{mehra1971}, the loss and gradient norm of test-time adaptation, and learned detectors trained on labelled shifted rollouts.

\item \emph{Test-time adaptation.} AdaJEPA \cite{wang2026adajepa}, OpTTA \cite{mete2026optta} and JEPA-TTT \cite{zhang2026jepattt} update model weights. Sandwich-Residuals \cite{soni2026sandwich} and ALRA \cite{soni2026alra} train residual modules.

\item \emph{Structural baselines.} Robust planning over the prior, an unstructured affine action map (ActLin), and the ablation Lie-SGD, fitting our Lie coordinates by gradient descent.
\end{itemize}

\paragraph{Implementation details.}
Unless stated otherwise, shifts are constant and present from the first block. They are drawn from a \emph{shift prior} over all $K$ coordinates, with rotations up to $\pm45^{\circ}$ and gains in $[0.5,1.25]$, and complemented by rotation--gain planes and by small shifts near the detection threshold (\cref{app:shifts}). Detection uses $124$, $70$ and $60$ shifts and recovery $30$, $20$ and $15$ prior draws on PushT, TwoRoom and Cube. The zero point $\thz$, the residual law $\Sigma_r$ and the over-dispersion $\kappa(m)$ are fitted on nominal rollouts before any shift is applied (\cref{app:calibration}). \actus pools the shift scores of fleets of $m=50$ episodes and runs five GN iterations per block. Once its alarm fires, it re-plans through $\Tf{\dhat}^{-1}$, so compensation enters the plan at block $\tau=1$ at the earliest.

\paragraph{Metrics.}
Detection is measured by the AUROC of a statistic on windows of $n$ transitions (one transition is one executed block) against nominal windows. Recovery is the success rate averaged over shifts drawn from the shift prior, with differences paired over the same draws. Estimation error is $\rho=\Fnorm{\dhat-\delta}/\Fnorm{\delta}$.

\subsection{Shift detection}
\label{sec:exp-detect}

\begin{figure}[t]
\centering
\includegraphics[width=\columnwidth]{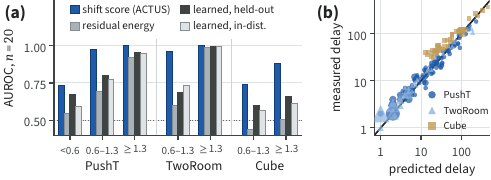}
\caption{\textbf{Shift detection.} \textbf{(a)} AUROC ($n=20$) by
Fisher norm of the shift. \textbf{(b)} Measured against predicted GLR
delay (\cref{eq:delay}).}
\label{fig:detect}
\end{figure}

\begin{table*}[t]
\centering
\caption{\textbf{Task success under actuation shifts.} Shared frozen
LeWM, planner, start--goal pairs, shift draws and replanning schedule.
$\Delta$: points against Frozen (shaded) and against \actus. Adapts: weights (W), residual modules (R). ms: update time per block before the alarm for $50$ episodes on A100, three seeds. P/C: PushT\,/\,Cube. SR: Success Rate. $^\dagger$Compensation at the scheduled replanning for ablation.}
\label{tab:main}
\footnotesize
\setlength{\tabcolsep}{2.6pt}
\renewcommand{\arraystretch}{0.94}
\begin{tabular}{@{}c@{\hspace{3pt}}l l c cccc cccc@{}}
\toprule
 & & & & \multicolumn{4}{c}{PushT} & \multicolumn{4}{c}{OGBench-Cube} \\
\cmidrule(lr){5-8}\cmidrule(l){9-12}
 & Method & Adapts\,{\tiny P/C} & time (ms)$\downarrow$\,{\tiny P/C} & SR$\uparrow$ & $\Delta$Frozen & $\Delta$\actus\ {\tiny[95\% CI]} & Nom.$\uparrow$ & SR$\uparrow$ & $\Delta$Frozen & $\Delta$\actus\ {\tiny[95\% CI]} & Nom.$\uparrow$ \\
\midrule
 & Frozen & -- & -- & 0.650 & -- & $-$21.2\,{\tiny\color{uwmgrayL}[$-$28.0,\,$-$14.9]} & 0.873 & 0.378 & -- & $-$5.4\,{\tiny\color{uwmgrayL}[$-$9.1,\,$-$1.9]} & 0.520 \\
\midrule
\multirow{5}{*}{\rotatebox[origin=c]{90}{\tiny\color{uwmgray}Test-time adapt.}} & AdaJEPA~\cite{wang2026adajepa} & W 2.6M & 49/51 & 0.627 & \cellcolor{uwmscarlet!7}$-$2.3 & $-$23.5\,{\tiny\color{uwmgrayL}[$-$29.1,\,$-$18.2]} & 0.847 & 0.347 & \cellcolor{uwmscarlet!13}$-$3.1 & $-$8.4\,{\tiny\color{uwmgrayL}[$-$12.8,\,$-$4.2]} & 0.485 \\
 & OpTTA~\cite{mete2026optta} & W 2.6M & 4997/5845 & 0.600 & \cellcolor{uwmscarlet!10}$-$5.0 & $-$26.2\,{\tiny\color{uwmgrayL}[$-$31.7,\,$-$21.0]} & 0.807 & 0.346 & \cellcolor{uwmscarlet!13}$-$3.2 & $-$8.6\,{\tiny\color{uwmgrayL}[$-$12.3,\,$-$4.6]} & 0.505 \\
 & JEPA-TTT~\cite{zhang2026jepattt} & W 11.7M & 103/52 & 0.673 & \cellcolor{uwmblue!8}$+$2.3 & $-$18.9\,{\tiny\color{uwmgrayL}[$-$23.8,\,$-$14.2]} & 0.940 & \underline{0.391} & \cellcolor{uwmblue!9}$+$1.3 & $-$4.1\,{\tiny\color{uwmgrayL}[$-$7.9,\,$-$0.2]} & 0.500 \\
 & Sandwich-Res.~\cite{soni2026sandwich} & R 87k & 52/58 & 0.634 & \cellcolor{uwmscarlet!6}$-$1.6 & $-$22.8\,{\tiny\color{uwmgrayL}[$-$28.7,\,$-$17.2]} & 0.847 & 0.368 & \cellcolor{uwmscarlet!7}$-$1.0 & $-$6.3\,{\tiny\color{uwmgrayL}[$-$10.4,\,$-$2.5]} & 0.485 \\
 & ALRA~\cite{soni2026alra} & R 37k & \underline{47/47} & 0.634 & \cellcolor{uwmscarlet!6}$-$1.6 & $-$22.8\,{\tiny\color{uwmgrayL}[$-$28.5,\,$-$17.4]} & 0.860 & 0.338 & \cellcolor{uwmscarlet!16}$-$4.0 & $-$9.4\,{\tiny\color{uwmgrayL}[$-$13.7,\,$-$4.5]} & 0.520 \\
\midrule
\multirow{2}{*}{\rotatebox[origin=c]{90}{\tiny\color{uwmgray}Structure}} & Robust planning~\cite{chua2018pets} & none & -- & 0.526 & \cellcolor{uwmscarlet!18}$-$12.4 & $-$33.6\,{\tiny\color{uwmgrayL}[$-$40.1,\,$-$27.3]} & 0.693 & 0.291 & \cellcolor{uwmscarlet!29}$-$8.7 & $-$14.0\,{\tiny\color{uwmgrayL}[$-$16.1,\,$-$12.0]} & 0.305 \\
 & {\color{uwmgray}\scriptsize abl. (Structure):}~ActLin & affine 110/650 & 142/155 & 0.778 & \cellcolor{uwmblue!24}$+$12.8 & $-$8.4\,{\tiny\color{uwmgrayL}[$-$11.6,\,$-$5.4]} & 0.907 & 0.364 & \cellcolor{uwmscarlet!8}$-$1.4 & $-$6.7\,{\tiny\color{uwmgrayL}[$-$10.5,\,$-$3.0]} & 0.510 \\
\midrule
 & {\color{uwmgray}\scriptsize abl.:}~Lie-SGD$^\dagger$ (Ours) & $\delta$ 4/6 & 508/635 & \underline{0.829} & \cellcolor{uwmblue!32}$+$17.9 & $-$3.3\,{\tiny\color{uwmgrayL}[$-$5.3,\,$-$1.3]} & 0.853 & 0.388 & \cellcolor{uwmblue!8}$+$1.0 & $-$4.4\,{\tiny\color{uwmgrayL}[$-$7.6,\,$-$1.2]} & 0.520 \\
\rowcolor{uwmblueXL}\multirow{-2}{*}{\rotatebox[origin=c]{90}{\tiny\color{uwmgray}Ours}} & \textbf{\actus (Ours)} & $\delta$ 4/6 & \textbf{43/47} & \textbf{0.862} & \cellcolor{uwmblue!38}$+$21.2 & -- & 0.873 & \textbf{0.432} & \cellcolor{uwmblue!26}$+$5.4 & -- & 0.520 \\
\midrule
\multirow{2}{*}{\rotatebox[origin=c]{90}{\tiny\color{uwmgray}Oracle}} & Oracle@1 & -- & -- & 0.864 & \cellcolor{uwmblue!38}$+$21.4 & $+$0.2\,{\tiny\color{uwmgrayL}[$-$0.7,\,$+$1.2]} & 0.893 & 0.440 & \cellcolor{uwmblue!29}$+$6.2 & $+$0.9\,{\tiny\color{uwmgrayL}[$-$0.7,\,$+$2.3]} & 0.485 \\
 & Oracle@0 & -- & -- & 0.854 & \cellcolor{uwmblue!36}$+$20.4 & $-$0.7\,{\tiny\color{uwmgrayL}[$-$1.7,\,$+$0.3]} & 0.873 & 0.467 & \cellcolor{uwmblue!40}$+$8.9 & $+$3.5\,{\tiny\color{uwmgrayL}[$+$1.6,\,$+$5.5]} & 0.520 \\
\midrule
\multirow{2}{*}{\rotatebox[origin=c]{90}{\tiny\color{uwmgray}Comp.}} & JEPA-TTT + \actus & W+$\delta$ & 144/99 & 0.877 & \cellcolor{uwmblue!40}$+$22.7 & $+$1.5\,{\tiny\color{uwmgrayL}[$+$0.6,\,$+$2.5]} & 0.867 & 0.431 & \cellcolor{uwmblue!25}$+$5.3 & $-$0.1\,{\tiny\color{uwmgrayL}[$-$1.5,\,$+$1.1]} & 0.520 \\
 & AdaJEPA (fleet) + \actus & W+$\delta$ & 69/80 & 0.877 & \cellcolor{uwmblue!40}$+$22.7 & $+$1.6\,{\tiny\color{uwmgrayL}[$+$0.7,\,$+$2.4]} & 0.873 & 0.434 & \cellcolor{uwmblue!27}$+$5.6 & $+$0.3\,{\tiny\color{uwmgrayL}[$-$1.2,\,$+$1.6]} & 0.505 \\
\bottomrule
\end{tabular}

\end{table*}

\begin{figure}[t]
\centering
\includegraphics[width=\columnwidth]{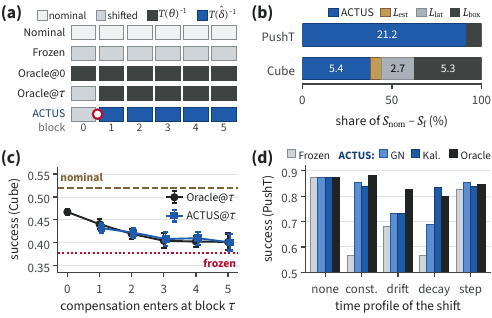}
\caption{\textbf{Recovery and its decomposition.} \textbf{(a)} What each
block is executed with; ring: alarm.
\textbf{(b)} Success lost to the shift, split into the part by
\actus and the three terms of Thm.~2. \textbf{(c)} Cube:
success against the block $\tau$ where compensation enters. \textbf{(d)}
Shifts constant, drifting, decaying or stepping at block $6$:
success of planners ($150$ episodes each); GN, Kal.: the two estimators of \actus.}
\label{fig:recovery}
\end{figure}

\begin{figure*}[t]
\centering
\includegraphics[width=\textwidth]{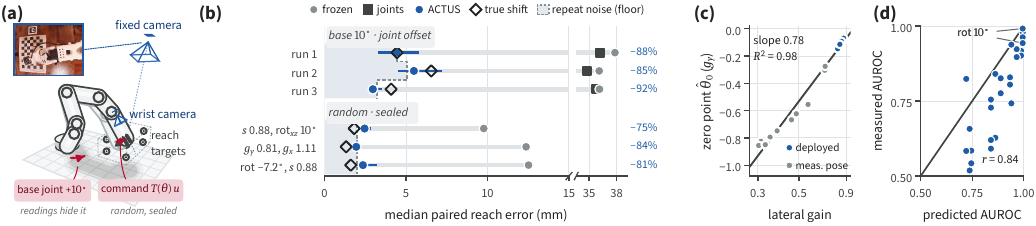}
\caption{\textbf{Real SO-101 arm.}
\textbf{(a)} Settings of the test environment.
\textbf{(b)} Median paired reach error for the frozen planner with \actus ($95\%$ bootstrap interval). Shaded: the error of
repeating the nominal reach unshifted.
\textbf{(c)} Zero point $\hat\theta_0(g_y)$ against the lateral gain
measured by the encoders, $21$ nominal blocks, two controllers.
\textbf{(d)} Measured against predicted AUROC at $n=20$, $13$
shifts $\times$ $2$ session blocks.}
\label{fig:real}
\end{figure*}

\paragraph{Detection accuracy and predictability.}
The shift score performs best in all seven settings of \cref{fig:detect}a, exceeding the strongest baseline by \(0.04\)--\(0.21\) AUROC except where all statistics saturate (\RQref{1}). On weak shifts, residual energy achieves only \(0.44\)--\(0.60\) AUROC in each weakest bin. The shift score instead uses the \(K\) sensitive directions \(J_t\) along which weak shifts move the residual, read directly from the frozen model. Even detectors trained on the test shifts themselves are competitive only on the strongest shifts, where any statistic separates (full comparisons in \cref{tab:detbase}).

Our theory predicts detection AUROC from non-centrality \(\lambda\), and all \(254\) shifts follow the predicted curves across generators, magnitudes, and tasks (\cref{fig:teaser}c). At \(n=20\), measured AUROC differs by at most \(0.05\) from the theoretical law (\cref{tab:detect}) and \(0.04\) from predictions using nominal windows alone (\cref{tab:supp-windows}). The fitted over-dispersion \(\kappa(m)\) keeps false alarms at the design level (\cref{app:calibration}).

\paragraph{Streaming detection.}
We run the GLR online on streams of shifted episodes at a $5\%$ false-alarm rate and record the transition at which it first fires. Detection delay tracks the prediction from accumulated information, with rank correlation at least $0.95$ in every environment (\cref{fig:detect}b). The same alarm triggers closed-loop compensation. \Cref{app:analysis} extends the comparison to shifts that vary within episode.

\subsection{Shift estimation and compensation}
\label{sec:exp-comp}
In this section, the five planners of \cref{fig:recovery}a give the success lost to the shift, $\Snom-\Sfro$, and measure each term of Thm.~2 separately: $\Lbox=\Snom-\Sor0$, $\Llat=\Sor0-\Sor\tau$ and $\Lest=\Sor\tau-\Sad$. 
We report PushT and Cube, since on TwoRoom the frozen planner already matches the nominal one ($0.752$ vs.\ $0.760$), leaving no success to recover (\cref{app:recovery-details}).

\paragraph{Comparison with baselines.}
The actuator shift costs the frozen planner $22.3$ points of success on PushT and $14.2$ on Cube. In \cref{tab:main}, \actus recovers $21.2$ and $5.4$, respectively, without changing model weights or nominal success (\RQref{2}). It matches Oracle@1, which receives the true shift at the same block, within statistical uncertainty, consistent with a negligible estimation contribution $\Lest$ in Thm.~2. The remaining Cube gap is mainly attributed to latency and action-box constraints (\cref{fig:recovery}b). In contrast, the five adaptation methods gain at most $2.3$ points over Frozen, four fall below it on both tasks, and nominal success drops by up to $6.6$ points. This may reflect actuator mismatch being absorbed into the predictor when fitting many parameters to few transitions. The structural baselines isolate the value of actuator modeling: robust planning without estimation loses $12.4$ points on PushT, an unstructured affine correction recovers $12.8$, and Lie-SGD with the same low-dimensional coordinates recovers $17.9$. \actus further improves recovery through calibrated detection and replanning at the alarm, which corrects the plan as soon as the shift is detected.
Model adaptation and actuator correction are complementary: run together with \actus and replanned at its alarm, JEPA-TTT and fleet AdaJEPA add $1.5$ and $1.6$ points to \actus on PushT and match it on Cube (see Composition rows). Under the adaptation methods' native protocol of thirty replannings per episode, adding \actus improves JEPA-TTT by $8.1$ points (\cref{tab:supp-native} provides details). 

\paragraph{Timing of compensation and ablations.}
In \cref{tab:main}, \actus compensates from the first block ($\tau=1$) in
$86$ of $90$ fleets on PushT and $43$ of $60$ on Cube. To measure the cost
of a later alarm, we delay compensation to block $\tau$ on Cube
(\cref{fig:recovery}c). Success drops from $0.432$ at $\tau=1$ to $0.410$ at
$\tau=4$, but stays within one point of Oracle@$\tau$ throughout. The loss
therefore comes from time spent without compensation rather than from
estimation, consistent with Thm.~2.
Two components of \actus set this timing (\cref{tab:supp-ablation}).
Replanning at the next scheduled block instead of at the alarm costs $3.4$
points on PushT and $3.1$ on Cube. Without the alarm gate, compensation
starts at block~$1$ everywhere: this gains less than one point under shift
but costs $7.5$ on nominal Cube, where the estimator reads a spurious shift.

\paragraph{Extended evaluations.}
For time-varying shifts on PushT (\cref{fig:recovery}d), the information Kalman filter of \actus tracks the changing actuator. On a decaying gain, it recovers $26.6$ points over Frozen, compared with $12.0$ for the cumulative GN estimate; on a constant shift, the two are comparable. On PushT, the alarm fires no later than the information-time prediction on every profile and never without shift (\cref{app:analysis}). Further experiments use the same weights and planner. Across $24$ Cube shifts persisting over fleet rounds, the carried estimate matches the same-round oracle from the second round onward, while a single robot running sequential episodes obtains $86\%$ of the oracle's gain within sixteen episodes (\cref{tab:persist}). The supplement also reports the full rotation--gain planes for PushT and Cube (\cref{fig:supp-pusht-grid,fig:cubegrid}), and identification baselines (\cref{tab:ident}).

\subsection{Shift attribution}
\label{sec:exp-attr}

Attribution asks which generator dominates, so it is evaluated on planes spanned by two generators (rotation--gain, $g_x$--$g_y$ and rotation--shear, \cref{app:shifts}). In \cref{tab:attr}, the rule of \cref{sec:attribute} names the correct generator $22/24$ on PushT and on TwoRoom and $15/16$ on Cube (\RQref{3}). The error bars of $\dhat$ are those of the theory: its bootstrap covariance on Cube is $1.25\times$ the Cram\'er--Rao bound. On the same planes, GN estimate behind this rule has the lowest error among schemes of \cref{app:identification}: a median $\rho$ of $0.07$ on PushT and $0.14$ on Cube, against at least $0.09$ and $0.25$ for the others (\cref{tab:ident}).

\subsection{Generalization to DINO world model}
\label{sec:exp-dino}
We repeat the PushT and Cube protocols on a second world model, DINO-WM \cite{zhou2025dinowm}, whose latent is a grid of frozen DINOv2 patch tokens instead of a single vector. All of the components are reused unchanged, and only the read-out space of the residual is model specific (\cref{app:dino}).

\actus transfers without retuning (\cref{tab:dino}, all sets in \cref{tab:supp-dino}), for both detection and compensation (\RQref{4}). Every shift raises the alarm, $104$ of $113$ at the first block, and no held-out nominal fleet does. 
On the prior draws \actus gains $18.0$ points over the
frozen planner on PushT and $7.0$ on Cube, level with Oracle@1,
with a median $\rho$ of $0.04$.

\begin{table}[t]
\centering
\caption{\textbf{\actus on a frozen DINO-WM.} In-plane: prior restricted
to rotation and gain. Alarmed: shifts raising the closed-loop alarm. False alarms on
held-out nominal fleets: \(0\%\).}
\label{tab:dino}
\footnotesize
\setlength{\tabcolsep}{2.4pt}
\begin{tabular}{@{}llcccccc@{}}
\toprule
 & Set & Alarmed & Frozen & \actus & Or.@1 & Or.@0 & Nom. \\
\midrule
PushT & prior (\(30\))          & 30/30 & 0.540 & 0.720 & 0.730 & 0.733 & 0.780 \\
      & plane (\(48\))          & 48/48 & 0.220 & 0.683 & 0.690 & 0.702 & 0.780 \\
      & in-plane (\(20\)) & 20/20 & 0.548 & 0.728 & 0.745 & 0.744 & 0.780 \\
Cube  & prior (\(15\))          & 15/15 & 0.359 & 0.429 & 0.422 & 0.453 & 0.500 \\
\bottomrule
\end{tabular}
\end{table}

\subsection{Real-world application}
\label{sec:exp-real}

\paragraph{Setup.}
We test \actus on a real robot arm, with a world model trained on the arm's own data. The arm is an SO-101 with a fixed and a wrist camera (\cref{fig:real}a). One LeWM per camera is trained on nominal data from this arm, frozen, and calibrated on nominal sessions. Shifts are injected in software on the commanded end-effector increment, or at the servo interface as a $10^\circ$ base-joint offset that the joint readings do not reveal. The task is an open-loop reach to one of eight targets, and the error of an episode is the distance between its end point under shift and that of the same seed without shift (\cref{app:real}).

\paragraph{Results.}
Detection on the arm follows the theory. Across $13$ shifts the measured window AUROC tracks the value predicted from nominal data ($r=0.84$, \cref{fig:real}d). \actus also sees a fault that the arm's own sensors miss. With the base joint $10^{\circ}$ off, the wrist camera reads a rotation of $10.1^{\circ}$ while the joint readings report none, and compensating with this estimate brings the median reach error from $36$--$38$\,mm down to $3.0$--$5.5$\,mm, as low as compensating with the true offset and level with the arm's repeat noise ($3.2$--$5.1$\,mm, the error of repeating the nominal reach; \cref{fig:real}b, top). Under three shifts drawn at random and sealed before the session, \actus obtains $92$--$94\%$ of the error reduction of the true shift ($p<10^{-11}$) and leaves $1.9$--$2.5$\,mm against a repeat noise of $2.0$\,mm (\cref{fig:real}b, bottom; \RQref{5}). The zero point read by \actus follows the arm's effective gain (\cref{fig:real}c), so each estimate is referenced to nominal blocks of its session, as Thm.~1(iii) prescribes (full results in \cref{app:real}).

\section{Conclusion}
\label{sec:conclusion}

A frozen world model answers what its actions
will do. We showed that it also answers the inverse question of which
actuation shift explains what it observed. \actus uses calibrated residuals to detect, estimate, attribute, and compensate shifts without weight updates, with performance known in advance from nominal data. We discuss the limitations of this work in \cref{app:limitations}.
We expect the inverse use of
world models to become as routine as their forward use.

\section*{Supplementary material}
Sections, algorithms, lemmas, propositions, tables and figures numbered
with the prefix S refer to the supplementary material, which will be
released upon acceptance.

{
    \small
    \bibliographystyle{ieeenat_fullname}
    \bibliography{main}

@article{rao1945,
  author    = {Rao, C. Radhakrishna},
  title     = {Information and the accuracy attainable in the estimation of statistical parameters},
  journal   = {Bulletin of the Calcutta Mathematical Society},
  volume    = {37},
  pages     = {81--91},
  year      = {1945}
}

@book{cramer1946,
  author    = {Cram{\'e}r, Harald},
  title     = {Mathematical Methods of Statistics},
  publisher = {Princeton University Press},
  year      = {1946}
}

@article{rao1948,
  author    = {Rao, C. Radhakrishna},
  title     = {Large sample tests of statistical hypotheses concerning several parameters with applications to problems of estimation},
  journal   = {Mathematical Proceedings of the Cambridge Philosophical Society},
  volume    = {44},
  number    = {1},
  pages     = {50--57},
  year      = {1948}
}

@article{wedderburn1974,
  author    = {Wedderburn, Robert W. M.},
  title     = {Quasi-likelihood functions, generalized linear models, and the {G}auss--{N}ewton method},
  journal   = {Biometrika},
  volume    = {61},
  number    = {3},
  pages     = {439--447},
  year      = {1974}
}

@book{lecam1986,
  author    = {Le Cam, Lucien},
  title     = {Asymptotic Methods in Statistical Decision Theory},
  publisher = {Springer},
  year      = {1986}
}

@book{lehmann2005,
  author    = {Lehmann, Erich L. and Romano, Joseph P.},
  title     = {Testing Statistical Hypotheses},
  edition   = {3rd},
  publisher = {Springer},
  year      = {2005}
}

@article{laird1982,
  author    = {Laird, Nan M. and Ware, James H.},
  title     = {Random-effects models for longitudinal data},
  journal   = {Biometrics},
  volume    = {38},
  number    = {4},
  pages     = {963--974},
  year      = {1982}
}

@article{lorden1971,
  author    = {Lorden, Gary},
  title     = {Procedures for reacting to a change in distribution},
  journal   = {The Annals of Mathematical Statistics},
  volume    = {42},
  number    = {6},
  pages     = {1897--1908},
  year      = {1971}
}

@book{basseville1993,
  author    = {Basseville, Mich{\`e}le and Nikiforov, Igor V.},
  title     = {Detection of Abrupt Changes: Theory and Application},
  publisher = {Prentice Hall},
  year      = {1993}
}

@article{lai1998,
  author    = {Lai, Tze Leung},
  title     = {Information bounds and quick detection of parameter changes in stochastic systems},
  journal   = {IEEE Transactions on Information Theory},
  volume    = {44},
  number    = {7},
  pages     = {2917--2929},
  year      = {1998}
}

@article{mehra1971,
  author    = {Mehra, Raman K. and Peschon, John},
  title     = {An innovations approach to fault detection and diagnosis in dynamic systems},
  journal   = {Automatica},
  volume    = {7},
  number    = {5},
  pages     = {637--640},
  year      = {1971}
}

@article{willsky1976,
  author    = {Willsky, Alan S.},
  title     = {A survey of design methods for failure detection in dynamic systems},
  journal   = {Automatica},
  volume    = {12},
  number    = {6},
  pages     = {601--611},
  year      = {1976}
}

@article{benveniste1987,
  author    = {Benveniste, Albert and Basseville, Mich{\`e}le and Moustakides, George V.},
  title     = {The asymptotic local approach to change detection and model validation},
  journal   = {IEEE Transactions on Automatic Control},
  volume    = {32},
  number    = {7},
  pages     = {583--592},
  year      = {1987}
}

@article{basseville1997,
  author    = {Basseville, Mich{\`e}le},
  title     = {Information criteria for residual generation and fault detection and isolation},
  journal   = {Automatica},
  volume    = {33},
  number    = {5},
  pages     = {783--803},
  year      = {1997}
}

@article{basseville1998,
  author    = {Basseville, Mich{\`e}le},
  title     = {On-board component fault detection and isolation using the statistical local approach},
  journal   = {Automatica},
  volume    = {34},
  number    = {11},
  pages     = {1391--1415},
  year      = {1998}
}

@article{basseville2002,
  author    = {Basseville, Mich{\`e}le and Nikiforov, Igor V.},
  title     = {Fault isolation for diagnosis: nuisance rejection and multiple hypotheses testing},
  journal   = {Annual Reviews in Control},
  volume    = {26},
  number    = {2},
  pages     = {189--202},
  year      = {2002}
}

@article{kalman1960,
  author    = {Kalman, Rudolf E.},
  title     = {A new approach to linear filtering and prediction problems},
  journal   = {Journal of Basic Engineering},
  volume    = {82},
  number    = {1},
  pages     = {35--45},
  year      = {1960}
}

@article{bell1993,
  author    = {Bell, Bradley M. and Cathey, Frederick W.},
  title     = {The iterated {K}alman filter update as a {G}auss--{N}ewton method},
  journal   = {IEEE Transactions on Automatic Control},
  volume    = {38},
  number    = {2},
  pages     = {294--297},
  year      = {1993}
}

@article{rubinstein1999,
  author    = {Rubinstein, Reuven Y.},
  title     = {The cross-entropy method for combinatorial and continuous optimization},
  journal   = {Methodology and Computing in Applied Probability},
  volume    = {1},
  number    = {2},
  pages     = {127--190},
  year      = {1999}
}

@book{nocedal2006,
  author    = {Nocedal, Jorge and Wright, Stephen J.},
  title     = {Numerical Optimization},
  edition   = {2nd},
  publisher = {Springer},
  year      = {2006}
}

@book{barfoot2017,
  author    = {Barfoot, Timothy D.},
  title     = {State Estimation for Robotics},
  publisher = {Cambridge University Press},
  year      = {2017}
}

@article{sola2018,
  author    = {Sol{\`a}, Joan and Deray, Jeremie and Atchuthan, Dinesh},
  title     = {A micro {L}ie theory for state estimation in robotics},
  journal   = {arXiv preprint arXiv:1812.01537},
  year      = {2018}
}

@book{pearl2009,
  author    = {Pearl, Judea},
  title     = {Causality: Models, Reasoning, and Inference},
  edition   = {2nd},
  publisher = {Cambridge University Press},
  year      = {2009}
}

@inproceedings{buesing2019woulda,
  author    = {Buesing, Lars and Weber, Th{\'e}ophane and Zwols, Yori and Racani{\`e}re, S{\'e}bastien and Guez, Arthur and Lespiau, Jean-Baptiste and Heess, Nicolas},
  title     = {Woulda, Coulda, Shoulda: Counterfactually-guided policy search},
  booktitle = {International Conference on Learning Representations},
  year      = {2019},
  note      = {arXiv:1811.06272}
}

@article{bear2023cwm,
  author    = {Bear, Daniel M. and Feigelis, Kevin and Chen, Honglin and Lee, Wanhee and Venkatesh, Rahul and Kotar, Klemen and Durango, Alex and Yamins, Daniel L. K.},
  title     = {Unifying (machine) vision via counterfactual world modeling},
  journal   = {arXiv preprint arXiv:2306.01828},
  year      = {2023}
}

@inproceedings{chua2018pets,
  author    = {Chua, Kurtland and Calandra, Roberto and McAllister, Rowan and Levine, Sergey},
  title     = {Deep reinforcement learning in a handful of trials using probabilistic dynamics models},
  booktitle = {Advances in Neural Information Processing Systems},
  year      = {2018}
}

@inproceedings{ha2018worldmodels,
  author    = {Ha, David and Schmidhuber, J{\"u}rgen},
  title     = {Recurrent world models facilitate policy evolution},
  booktitle = {Advances in Neural Information Processing Systems},
  year      = {2018}
}

@inproceedings{hafner2019planet,
  author    = {Hafner, Danijar and Lillicrap, Timothy and Fischer, Ian and Villegas, Ruben and Ha, David and Lee, Honglak and Davidson, James},
  title     = {Learning latent dynamics for planning from pixels},
  booktitle = {International Conference on Machine Learning},
  series    = {PMLR},
  volume    = {97},
  pages     = {2555--2565},
  year      = {2019}
}

@inproceedings{hafner2020dreamer,
  author    = {Hafner, Danijar and Lillicrap, Timothy and Ba, Jimmy and Norouzi, Mohammad},
  title     = {Dream to control: Learning behaviors by latent imagination},
  booktitle = {International Conference on Learning Representations},
  year      = {2020}
}

@article{assran2025vjepa2,
  author    = {Assran, Mahmoud and Bardes, Adrien and Fan, David and Garrido, Quentin and Howes, Russell and Komeili, Mojtaba and Muckley, Matthew and Rizvi, Ammar and Roberts, Claire and Sinha, Koustuv and others},
  title     = {{V-JEPA 2}: Self-supervised video models enable understanding, prediction and planning},
  journal   = {arXiv preprint arXiv:2506.09985},
  year      = {2025}
}

@inproceedings{sobal2025pldm,
  author    = {Sobal, Vlad and Zhang, Wancong and Cho, Kyunghyun and Balestriero, Randall and Rudner, Tim G. J. and LeCun, Yann},
  title     = {Learning from reward-free offline data: A case for planning with latent dynamics models},
  booktitle = {Advances in Neural Information Processing Systems},
  year      = {2025},
  note      = {arXiv:2502.14819}
}

@inproceedings{zhou2025dinowm,
  author    = {Zhou, Gaoyue and Pan, Hengkai and LeCun, Yann and Pinto, Lerrel},
  title     = {{DINO-WM}: World models on pre-trained visual features enable zero-shot planning},
  booktitle = {International Conference on Machine Learning},
  series    = {PMLR},
  volume    = {267},
  pages     = {79115--79135},
  year      = {2025},
  note      = {arXiv:2411.04983}
}

@article{maes2026lewm,
  author    = {Maes, Lucas and Le Lidec, Quentin and Scieur, Damien and LeCun, Yann and Balestriero, Randall},
  title     = {{LeWorldModel}: Stable end-to-end joint-embedding predictive architecture from pixels},
  journal   = {arXiv preprint arXiv:2603.19312},
  year      = {2026}
}

@article{maes2026swm,
  author    = {Maes, Lucas and Le Lidec, Quentin and Haramati, Dan and Massaudi, Nassim and Scieur, Damien and LeCun, Yann and Balestriero, Randall},
  title     = {stable-worldmodel-v1: Reproducible world modeling research and evaluation},
  journal   = {arXiv preprint arXiv:2602.08968},
  year      = {2026}
}

@inproceedings{nam2026,
  author    = {Nam, Heejeong and Le Lidec, Quentin and Maes, Lucas and LeCun, Yann and Balestriero, Randall},
  title     = {Causal-{JEPA}: Learning world models through object-level latent masking},
  booktitle = {International Conference on Machine Learning},
  series    = {PMLR},
  volume    = {306},
  pages     = {91808--91832},
  year      = {2026},
  note      = {arXiv:2602.11389}
}

@inproceedings{vasilev2026geometry,
  author    = {Vasilev, Vasil and Nayak, Siddharth},
  title     = {Does {LeWorldModel} preserve local action geometry?},
  booktitle = {NeurIPS Workshop on Symmetry and Geometry in Neural Representations},
  year      = {2026},
  note      = {arXiv:2610.31716}
}

@article{wang2026groupactions,
  author    = {Wang, Zijie and Zhang, Wei and Zhang, Weiming and Zhang, Fanqi and Tan, Xiao and Qin, Yipeng and Li, Guanbin},
  title     = {World models as group actions},
  journal   = {arXiv preprint arXiv:2605.24578},
  year      = {2026}
}

@inproceedings{chi2023diffusionpolicy,
  author    = {Chi, Cheng and Feng, Siyuan and Du, Yilun and Xu, Zhenjia and Cousineau, Eric and Burchfiel, Benjamin and Song, Shuran},
  title     = {Diffusion policy: Visuomotor policy learning via action diffusion},
  booktitle = {Robotics: Science and Systems},
  year      = {2023}
}

@inproceedings{park2025ogbench,
  author    = {Park, Seohong and Frans, Kevin and Eysenbach, Benjamin and Levine, Sergey},
  title     = {{OGBench}: Benchmarking offline goal-conditioned {RL}},
  booktitle = {International Conference on Learning Representations},
  year      = {2025}
}

@inproceedings{nagabandi2019adapt,
  author    = {Nagabandi, Anusha and Clavera, Ignasi and Liu, Simin and Fearing, Ronald S. and Abbeel, Pieter and Levine, Sergey and Finn, Chelsea},
  title     = {Learning to adapt in dynamic, real-world environments through meta-reinforcement learning},
  booktitle = {International Conference on Learning Representations},
  year      = {2019},
  note      = {arXiv:1803.11347}
}

@inproceedings{sun2020ttt,
  author    = {Sun, Yu and Wang, Xiaolong and Liu, Zhuang and Miller, John and Efros, Alexei A. and Hardt, Moritz},
  title     = {Test-time training with self-supervision for generalization under distribution shifts},
  booktitle = {International Conference on Machine Learning},
  series    = {PMLR},
  volume    = {119},
  pages     = {9229--9248},
  year      = {2020}
}

@inproceedings{wang2021tent,
  author    = {Wang, Dequan and Shelhamer, Evan and Liu, Shaoteng and Olshausen, Bruno and Darrell, Trevor},
  title     = {Tent: Fully test-time adaptation by entropy minimization},
  booktitle = {International Conference on Learning Representations},
  year      = {2021}
}

@article{mete2026optta,
  author    = {Mete, Akshay and Kalathil, Dileep and Kumar, P. R.},
  title     = {Adapt for planning, not just prediction: Optimistic test-time world model adaptation},
  journal   = {arXiv preprint arXiv:2610.35636},
  year      = {2026}
}

@inproceedings{soni2026alra,
  author    = {Soni, Krishnam and Sehgal, Aditya and Dave, Vedant and Rueckert, Elmar},
  title     = {{ALRA}: Action-latent residual adaptation for world models},
  booktitle = {NeurIPS Workshop on PTA: From Pretrained Representations to Acting Agents},
  year      = {2026}
}

@article{soni2026sandwich,
  author    = {Soni, Krishnam and Sehgal, Aditya and Dave, Vedant and Rueckert, Elmar},
  title     = {Sandwich-residuals: Parameter-efficient test-time adaptation of world models},
  journal   = {arXiv preprint arXiv:2609.21740},
  year      = {2026}
}

@article{wang2026adajepa,
  author    = {Wang, Ying and Bounou, Oumayma and LeCun, Yann and Ren, Mengye},
  title     = {{AdaJEPA}: An adaptive latent world model},
  journal   = {arXiv preprint arXiv:2606.32026},
  year      = {2026}
}

@article{zhang2026jepattt,
  author    = {Zhang, Zheyuan and Ye, Suyu and Agarwal, Nakul and Nourkhiz Mahjoub, Hossein and Moradi Pari, Ehsan and Khashabi, Daniel and Shu, Tianmin and Tadiparthi, Vaishnav},
  title     = {{JEPA-TTT}: Persistent test-time training of latent world models for planning under dynamics shifts},
  journal   = {arXiv preprint arXiv:2610.00722},
  year      = {2026}
}

@inproceedings{domberg2025wmanomaly,
  author    = {Domberg, Fabian and Schildbach, Georg},
  title     = {World Models for Anomaly Detection during Model-Based Reinforcement Learning Inference},
  booktitle = {IEEE/RSJ International Conference on Intelligent Robots and Systems},
  pages     = {4182--4187},
  year      = {2025},
  note      = {arXiv:2503.02552}
}
}

\end{document}